\documentclass{article}

\usepackage[final]{best_roai_thesis_awards_2025}
\usepackage{graphicx}
\usepackage[utf8]{inputenc} 
\usepackage[T1]{fontenc}    
\usepackage{hyperref}       
\usepackage{url}            
\usepackage{booktabs}       
\usepackage{amsmath}        
\usepackage{amsfonts}       
\usepackage{amssymb}
\usepackage{bbm}
\usepackage{nicefrac}       
\usepackage{microtype}      
\usepackage{xcolor}         
\usepackage[numbers]{natbib}

\title{Semi-Supervised Learning for Large Language Models Safety and Content Moderation}

\author{%
  Eduard Ștefan Dinuță$^1$, Iustin Sîrbu$^{1,2}$, Traian Rebedea$^{1,3}$ \\
  $^1$National University of Science and Technology Politehnica Bucharest \\
  $^2$Renius Technologies, $^3$NVIDIA \\
  \texttt{eduarddinuta3@gmail.com, iustin.sirbu@upb.ro, traian.rebedea@upb.ro} \\
}

\begin{document}

\maketitle

\begin{abstract}
   Safety for Large Language Models (LLMs) has been an ongoing research focus since their emergence and is even more relevant nowadays with the increasing capacity of those models. Currently, there are several guardrails in place for all public LLMs and multiple proposed datasets for training safety classifiers. 
   However, training these safety classifiers relies on large quantities of labeled data, which can be problematic to acquire, prone to labeling errors, or often include synthetic data.
   To address these issues, we suggest a different approach: utilizing semi-supervised learning techniques, which leverage both labeled and unlabeled data, to improve the performance on the safety task. We analyze the improvements that these techniques can offer for both prompts given to Large Language Models and the responses to those requests. Moreover, since augmentation is the central part of semi-supervised algorithms, we demonstrate the importance of using task-specific augmentations, which significantly increase the performance when compared to general-purpose augmentation techniques.
\end{abstract}

\section{Introduction}

Currently, all commercial Large Language Models (LLMs) have some kind of guardrails against malicious prompts and responses, either by moderating the model itself or adding another system on top of it that deals with the safety aspect. The problem is that this approach suffers from the same shortcomings as many other Machine Learning systems: gathering high-quality annotated data, which the models depend on. In general, data annotation is a cumbersome and costly task and can also lead to inaccuracies, either due to the subjective nature of the task or errors coming from deferring it to AI models. To solve this issue, we propose using semi-supervised learning (SSL). This approach leverages a small labeled dataset alongside a vast amount of unlabeled data from the real world, achieving better results at a reduced annotation cost. Building on this idea, this paper represents a starting point with two main contributions: 

    \begin{itemize}
        \item We perform an analysis on several state-of-the-art semi-supervised learning algorithms in the context of LLM safety. We focus on two categories: prompt harmfulness and response harmfulness, as we want to ensure that the models do not comply with malicious requests, but also do not offer harmful replies to benign questions. 
        \item We introduce a new, task-specific augmentation technique and show that it can significantly improve performance by doing a comparison between classic augmentation methods such as backtranslation and our custom LLM-generated augmentations that focus on the safety aspect.
    \end{itemize}

\section{Related Work}
    Multiple studies have been conducted in the field of LLM safety, to create either high-quality datasets and well-defined risk categories, or innovative training methods and models that are more capable at resisting a large set of attacks.
    
    WildGuard \cite{han2024wildguardopenonestopmoderation} introduces both a model and a large train dataset of 86,759 samples in 13 risk subcategories to act as an LLM moderation tool with large coverage, especially for adversarial attacks that have proven to be a serious problem. The test set contains 5K high-quality human-annotated examples covering broad risk scenarios. Aegis 2.0 \cite{ghosh2025aegis20diverseaisafety} proposes a taxonomy to classify safety risks into 12 top-level categories and 9 fine-grained ones. The Aegis dataset is fully open source and commercial and does not rely on synthetic data, providing 34K samples of human-LLM interactions carefully annotated by both humans and an LLM jury. While $85\%$ of the WildGuard training set is generated using GPT, Aegis collects its data from public sources, which makes it a suitable dataset for commercial use.

    When it comes to safety training techniques, multiple approaches have been proposed. One first category involves inducing safety-related behaviors directly into the model's weights, during training, a process known as alignment. One popular technique is
    Reinforcement Learning from Human Feedback (RLFH) \cite{NEURIPS2022_b1efde53} which combines a reward system with human evaluation. However, even after alignment, many LLMs remain prone to respond to unsafe queries, hence post-training methods have been developed. Some of these methods are fine-tuning a dedicated safety classifier, such as Llama Guard \cite{inan2023llamaguardllmbasedinputoutput} or fully programable frameworks, such as NeMo Guardrails \cite{rebedea2023nemo}. Other approaches include ensemble methods, such as the multi-judge architecture introduced in Aegis 2.0 \cite{ghosh2025aegis20diverseaisafety} and safety agents such as GuardAgent \cite{xiang2025guardagentsafeguardllmagents}, that get access to specifications, logs, and tools to enforce safety rules and constantly monitor and filter interactions.
         
\section{Approach}   
    \subsection{Baseline model}
    We initially experimented with several pretrained language models. The standard \emph{BERT-base-uncased} \cite{DBLP:journals/corr/abs-1810-04805} model lacked sufficient representational capacity to capture the nuances required in classifying LLM prompts and responses as harmful or safe. We also evaluated \emph{hateBERT} \cite{caselli-etal-2021-hatebert}, which is specifically trained on toxic online posts, but its training resulted in a limited generalization to other types of harmful content from the variety of classes in our dataset. In the end, we selected \emph{microsoft/deberta-v3-base} \cite{he2021debertav3} \cite{he2021deberta} as our primary model and tokenizer, as it consistently produced higher-quality embeddings and better results for our task.
    Moreover, because it creates separate embeddings for words and position, it does not have a token limit, so we were able to keep the same setup for the response classification task as well, which included longer texts. We applied supervised training on 200 and 2000 examples, as well as the entire dataset to test the limit of our model. Semi-supervised approaches are also tested on 200 and 2000 samples for comparison. During the early stages of our research, we also evaluated performance on 20K labeled samples, but did not obtain significant improvements. We concluded that our model reaches a plateau at around 2000 samples, which is further confirmed by the similar performance of models trained on that amount of samples compared to the entire dataset, presented in the results section.
    \subsection{Semi-supervised learning}

    To take advantage of the benefits of unlabeled data, we first applied Fixmatch \cite{sohn2020fixmatch}, which combines pseudo-labeling and consistency regularization. The total loss $l$ is computed as the sum of two loss terms  $l = l_{s} + \lambda _{u}l_{u}$ where $\lambda _{u}$ is a weighting parameter, $l_{s}$ is the standard cross-entropy loss on the labeled data, and $l_{u}$ is the loss on unlabeled data. The unlabeled term is computed as the cross-entropy loss between the hard pseudo-labels and the model's predictions on strongly augmentated inputs, only for the samples that exceed the confidence threshold: 

\vspace{-3mm}
\begin{equation}
    l_{u} = \frac{1}{\mu B} \sum_{b=1}^{B} \mathbbm{1}_{\tau} (q_{b})H(\hat{q_{b}}, p_{m}(\mathcal{A}(u_{b})))
\end{equation}

where $\mu$ is the ratio of unlabeled to labeled examples is a batch, \emph{B} is the batch size, ${q_{b}}$ is the probability distribution over the classes predicted on the weakly augmented input,  $\mathbbm{1}_{\tau}$ is used to filter out examples based on the confidence threshold $\tau$, $\displaystyle \hat{q_{b}} = \arg\max_y(q_{b})$ is the pseudo-label, $p_{m}$ is the model's current prediction, and $\mathcal{A}(u_{b})$ is the hard augmentation of the current example.

Next, we tested MarginMatch \cite{sosea2023marginmatch}, which adds an additional filter when choosing which examples are included in the unlabeled loss. This filter is based on the Average Pseudo Margin (APM), an adaptation of the Area Under Margin (AUM)~\cite{pleiss2020identifying} for pseudo-labels. First, a Pseudo-Margin (PM) is calculated for every sample and every class \emph{c} after each epoch, and it is equal to the difference between the logit corresponding to class \emph{c} and the highest other logit:

\begin{equation}
    PM^{(t)}_{c} = z^{(t)}_{c} - \max_{i \ne c} z^{(t)}_{i}
\end{equation}

Then, the APM for a sample $x$ and class $c$ is defined as the average of PMs from the previous epochs:

\begin{equation}
    APM^{(t)}_{c}(x) = \frac{1}{t}\sum_{i=1}^{t}PM^{(t)}_{c}(x)
\end{equation}

This APM formula is further adapted using an exponential moving average:

\begin{equation}
    \displaystyle
    APM^{(t)}_{c}(x)
    = PM^{(t)}_{c}(x)\,\frac{\delta}{t + 1}
      + APM^{(t - 1)}_{c}(x)\,\Bigl(1 - \frac{\delta}{t + 1}\Bigr)
\end{equation}

The APM is used as an additional confidence filter for incorporating examples into the unlabeled loss term. A sample is retained only if the APM exceeds a certain threshold, ensuring that the pseudo-label is historically stable and not merely an error of the current iteration.

Last but not least, we performed experiments with MultiMatch \cite{sirbu2025multimatch}, which offers a holistic approach that combines multiple classification heads, adaptive thresholds, and historical predictions. In our experiments, we used three classification heads as that was the best value reported in the original paper. The supervised loss term is calculated normally, only summed for every head with differences arising for the unsupervised loss. The classic head agreement of a multihead architecture is replaced with a Pseudo-Label Weighting Module (PLWM) which also takes into account disagreements, if one head is confident enough in its prediction. For this algorithm, the weight of a sample, generated by the PLWM for a specific head can be defined as:

\vspace{-3mm}
\begin{equation}
    W_{Multi}^{h} = [1 \cdot (\mathbbm{1}_{Multi}^{i} \wedge \mathbbm{1}_{Multi}^{i} \wedge \mathbbm{1}_{Agree}^{i}) + w_d \cdot (\mathbbm{1}_{Multi}^{i} \oplus \mathbbm{1}_{Multi}^{i})] \cdot \mathbbm{1}_{FreeMulti}^{h}
\end{equation}

where $\mathbbm{1}_{FreeMulti}^{h} = \mathbbm{1}_{Free}^{i} \vee \mathbbm{1}_{Free}^{j}$ is the union of the adaptive thresholds $\mathbbm{1}_{Free}^{k} = \mathbbm{1}(\max(q_b) > \tau_t(\hat{q}_b))$ for the remaining two heads and $\tau$ the current adaptive threshold for that class. $\mathbbm{1}_{Agree}^h = [\hat{q}_{b}^{i} = \hat{q}_{b}^{j}], \{i, j\} = \{1, 2, 3\} \setminus \{h\}$ is the function that checks whether the two remaining heads agree on the pseudo-label, and $\mathbbm{1}_{Multi}^{k} = \mathbbm{1}(APM_{\hat{q}_b^{k}}^{(t)}(u_b) > \gamma_{k, c}^{t - 1})$ is the APM filter used in MarginMatch adapted to multiple heads, with a dynamic threshold $\gamma_{k, c}^{t - 1}$ calculated as a percentile of examples where an agreement between heads has been reached. This way, MultiMatch also considers disagreements that can provide valuable information. The first term of the sum in parentheses represents the \textit{useful \& easy} examples, which are the ones where both heads agreed on the label, and the second term the \textit{useful \& hard} ones which are samples where a head provided a label, but the other one did not pass its APM confidence threshold. 

Finally, the unsupervised loss is computed for every head and summed using the resulting weights for each example, with the sample being excluded if the weight is zero:

\vspace{-5mm}
\begin{equation}
    l_{u, t} = \displaystyle\sum_{h = 1}^{3}\frac{1}{\mu B} \displaystyle\sum_{b = 1}^{\mu B} W_{Multi}^{h} \cdot H(\hat{q}_{b}^{h}, p_{m}(\mathcal{A}(u_{b}))
\end{equation}

\vspace{-4mm}
\section{Experimental Analysis}
\vspace{-1mm}
   We used WildGuardMix \cite{han2024wildguardopenonestopmoderation} for our training and test set, as it is a comprehensive set with a high coverage of safety categories and a focus on adversarial examples. From the train dataset, we filtered out the texts that were not in English, empty texts, samples with missing labels, and also the texts that were made up of special characters or did not represent a request, such as mathematical equations. We then performed a stratified sampling to obtain a train set of size 77,013 and a validation set of size 8557 for prompt classification. For the response task, we obtained a training set of 34,038 prompt-response pairs and a validation set of 3784 pairs with the same previous process. For our test set, we used the \emph{WildGuardTest} split, which resulted in 1696 examples after applying the same filters. While our model is trained only on WildGuard, we also evaluated its performance on three other datasets: OAIMod \cite{openai2022moderation}, XSTest \cite{rottger2023xstest} and Aegis 2.0 \cite{ghosh2025aegis20diverseaisafety}, to assess its generalization capabilities.

  We sampled subsets of 200 and 2000 labeled examples from the training set, with the rest being used as unlabeled data. The labeled set is randomly selected from both classes (harmful/unharmful) in equal numbers. To obtain significant results, we selected the labeled data using 3 different seeds and reported the average of the results on those seeds. The reported metric is \emph{harmfulness F1}, or the F1 score on the harmful (positive) class, as this is the metric commonly used on the topic of safety. 

   An essential part of this paper is to show how our proposed task-specific augmentation technique can improve the scores obtained on the safety task. As labeling a text as safe or harmful is a nuanced task and unsafe prompts are usually masked, with only some small fragments being the actual malicious intent, we claim that a deeper understanding of the given text is needed. For that, we prompted LLMs to identify the words or phrases in the given texts that might have a harmful connotation and replace them with alternatives that keep the same malicious intent, while also lightly paraphrasing the rest of the text to increase variance. We employed two uncensored models to generate augmentations for both safe and harmful prompts and their corresponding responses: 
   \emph{huihui-ai/Llama-3.2-3B-Instruct-abliterated} \cite{huihuiai2024llama3}  and \emph{Mistral-7B-Instruct-v0.1} \cite{jiang2023mistral7b}. For every sample, we generated one augmentation with each model, then randomly selected one of them during the training process.
   To evaluate the effectiveness of our method, we compared it with backtranslation, a standard technique widely used for text augmentations. Text is translated from English to an intermediate language and then back in English, the variance resulting from the differences that might appear in translations. We used \emph{OpusMT} \cite{TiedemannThottingal:EAMT2020} as our translation model and created 4 augmentations for every example in the following languages: Russian, German, French, and Romanian. However, these augmentations proved to not be good enough for our task, with them being either too simple or completely changing the text for languages like Russian. The results are summarized in Table~\ref{results_table}.

\vspace{-3mm}
\begin{table*}
  \centering
  \resizebox{\textwidth}{!}{%
  \begin{tabular}{lllccc|ccc}
    \toprule
    \textbf{Algorithm} & \textbf{Number Labeled} & \textbf{Augmentation} 
    & \multicolumn{3}{c|}{\textbf{Prompt Classification (F1)}} 
    & \multicolumn{3}{c}{\textbf{Response Classification (F1)}} \\
    \cmidrule(lr){4-6} \cmidrule(lr){7-9}
    & & & \textbf{WildGuard} & \textbf{OAIMod} & \textbf{Aegis 2.0} 
          & \textbf{WildGuard} & \textbf{XSTest} & \textbf{Aegis 2.0} \\
    \midrule
    Supervised & 200 & - & 0.748 & 0.531 & 0.642 & 0.586 & 0.560 & 0.566 \\
    FixMatch & 200 & Backtranslation & 0.746 & 0.544 & 0.546 & \textbf{0.610} & \textbf{0.630} & \textbf{0.640} \\
    MultiMatch & 200 & Backtranslation & 0.780 & 0.547 & 0.571 & \textbf{0.611} & 0.577 & 0.544\\
    MarginMatch & 200 & Backtranslation & 0.767 & 0.518 & 0.570 & \textbf{0.611} & 0.482 & 0.547\\
    FixMatch & 200 & LLM (our approach) & 0.783 & 0.544 & 0.670 & 0.604 & 0.591 & 0.571 \\
    MultiMatch & 200 & LLM (our approach) & \textbf{0.795} & \textbf{0.589} & \textbf{0.672} & 0.607 & 0.563 & 0.538 \\ 
    MarginMatch & 200 & LLM (our approach) & 0.783 & 0.583 & 0.670 & 0.588 & 0.469 & 0.505\\
    \midrule
    Supervised & 2000 & - & 0.828 & \textbf{0.659} & 0.769 & 0.689 & 0.641 & 0.632\\
    FixMatch & 2000 & Backtranslation & 0.846 & 0.628 & 0.749 & 0.693 & 0.550 & 0.621\\
    MultiMatch & 2000 & Backtranslation & 0.840 & 0.630 & 0.742 & \textbf{0.709} & 0.557 & 0.621\\
    MarginMatch & 2000 & Backtranslation & 0.836 & 0.627 & 0.725 & 0.686 & 0.533 & 0.515\\
    FixMatch & 2000 & LLM (our approach) & \textbf{0.854} & 0.595 & \textbf{0.776} & 0.700 & 0.595 & 0.636 \\
    MultiMatch & 2000 & LLM (our approach) & \textbf{0.856} & 0.585 & \textbf{0.776} & 0.704 & \textbf{0.672} & \textbf{0.683} \\
    MarginMatch & 2000 & LLM (our approach) & \textbf{0.855} & 0.579 & 0.762 & 0.691 & 0.572 & 0.641\\
    \midrule
    FullySupervised & 77K/34K & - & 0.870 & - & - & 0.721 & - & -\\
    \bottomrule
  \end{tabular}
  }
  \caption{\label{results_table}
    Results on prompt and response classification tasks, with LLM augmentations being the ones introduced by us. Best results for each subtask are in \textbf{bold}.
  }
  \vspace{-3mm}
\end{table*}

\section{Discussion}
\vspace{-2mm}

    Table~\ref{results_table} shows that semi-supervised learning significantly outperforms the supervised approach on the majority of benchmarks. Notably, in low data regimes with only 200 labeled examples, SSL approaches employing our proposed augmentation method provide improvements of $4\%-5\%$ across all datasets for prompt classification. Another impressive observation is that with only 2000 labeled examples we achieve a score of $85.6\%$ on WildGuard \cite{han2024wildguardopenonestopmoderation}, only $1.4\%$ behind the FullySupervised approach using the entire dataset. For OAIMod \cite{openai2022moderation} semi-supervised learning does not seem to help when the number of data is increased. That might be correlated with two limitations of our research. First of all, we are using a smaller model with 180 million parameters compared to the LLMs that are usually fine-tuned for this task, so generalization can be weaker. Secondly, we acknowledge that the unlabeled set must be diverse to cover multiple distributions. During our training, we sampled labeled and unlabeled examples from the same dataset, which is a process that can be improved by integrating multiple unlabeled data sources for better generalization. For response classification, we once again have better results when semi-supervised learning is employed, especially on other datasets and a small number of examples. For both XSTest \cite{rottger2023xstest} and Aegis 2.0 \cite{ghosh2025aegis20diverseaisafety} we increased the F1 score by $7\%-8\%$ and with 2000 examples we again have improvements on all datasets and get very close to the performance on the entire labeled dataset with a difference of only $1.2\%$.

    We must also highlight the importance of our LLM-generated augmentations, that succeed in capturing the nuanced, malicious intent of harmful prompts, which is often lost when using task-agnostic techniques. For prompt classification, there is a significant difference between backtranslation and the custom LLM augmentations, especially for 200 examples. Models with LLM augmentations have a significantly higher score compared to their backtranslation alternatives, with differences as large as $10\%$ for Aegis 2.0 \cite{ghosh2025aegis20diverseaisafety}. When it comes to response classification, an interesting observation is that for a small number of examples, backtranslation performs better. This can be attributed to the quality of the augmentations. The smaller LLMs that we used for this task sometimes struggled to paraphrase the given text, producing noisy outputs that affected performance for a small number of samples. However, for 2000 examples, LLM augmentation is more stable and highly outperforms backtranslation, with differences as high as $11.5\%$.

\vspace{-2mm}
\section{Conclusion}
\vspace{-2mm}
    As there are no other studies on the application of semi-supervised learning for the safety task that we are aware of, this research represents a starting point for further investigation. We successfully demonstrated that SSL is highly suitable for LLM safety classification, leading to enhanced performance compared to the supervised approach. Furthermore, we introduced a task-specific LLM-based augmentation technique that outperforms classic backtranslation on the majority of benchmarks, confirming the need for nuance-aware data augmentation in LLM safety research. 
    
\begin{ack}
This research was supported by the project “Romanian Hub for Artificial Intelligence - HRIA”, Smart Growth, Digitization and Financial Instruments Program, 2021-2027, MySMIS no. 351416.
\end{ack}

\small{
\bibliography{main}
\bibliographystyle{plainnat}
}

\end{document}